\documentclass[runningheads]{llncs}
\usepackage[T1]{fontenc}
\usepackage{graphicx,verbatim}
\usepackage{amsmath}
\usepackage{amsfonts}
\usepackage{dsfont}
\usepackage{graphicx}
\usepackage{tikz}
\usepackage{multirow}
\usepackage{array}
\usepackage{tabularx,booktabs}
\usepackage{subcaption}
\usepackage{dirtytalk}
\usepackage{bm}
\usepackage{dsfont}
\usepackage{float}
\usetikzlibrary{arrows.meta}
\usepackage{color}
\usepackage{hyperref}
\usepackage{rotating}
\usepackage{mathrsfs}
\usepackage[misc]{ifsym}

\definecolor{cgbla}{rgb}{0.5, 0, 0.5}
\definecolor{cliver}{rgb}{0.2, 0.8, 0.85}
\definecolor{cgrasper}{rgb}{0.2, 0.85, 0.3}

\newcommand{\gbla}[1]{{\color{cgbla}{#1}}}
\newcommand{\liver}[1]{{\color{cliver}{#1}}}
\newcommand{\grasper}[1]{{\color{cgrasper}{#1}}}

\begin{document}
\title{Revisiting the Evaluation Bias Introduced\\by Frame Sampling Strategies in\\Surgical Video Segmentation Using SAM2}

\titlerunning{Revisiting the Evaluation Bias Introduced by Frame Sampling Strategies}

\author{Utku Ozbulak$^{1,2}$(\Letter),
Seyed Amir Mousavi$^{1,2}$,\\
Francesca Tozzi$^{3,4}$,
Niki Rashidian$^{4,5}$, 
Wouter Willaert$^{3,4}$,\\
Wesley De Neve$^{1,2}$, and
Joris Vankerschaver$^{1,6}$
}

\authorrunning{Ozbulak et al.}
\institute{
$^{1}$Center for Biosystems and Biotech Data Science, Ghent University Global Campus, Incheon, Republic of Korea\\
$^{2}$IDLab, ELIS, Ghent University, Ghent, Belgium \\
$^{3}$Department of GI Surgery, Ghent University Hospital, Ghent, Belgium\\
$^{4}$Department of Human Structure and Repair, Ghent University, Ghent, Belgium\\
$^{5}$Department of HPB Surgery \& Liver Transplantation,\\ Ghent University Hospital, Ghent, Belgium\\
$^{6}$Department of Mathematics, Computer Science and Statistics,\\Ghent University, Ghent, Belgium \\
(\Letter) \email{utku.ozbulak@ghent.ac.kr}
}

\maketitle              
\begin{abstract}
\let\thefootnote\relax\footnotetext{Accepted for publication in the 28th International Conference on Medical Image Computing and Computer Assisted Intervention (MICCAI) Workshop on Fairness of AI in Medical Imaging (FAIMI), 2025.}
Real-time video segmentation is a promising opportunity for AI-assisted surgery, offering intraoperative guidance by identifying tools and anatomical structures. Despite growing interest in surgical video segmentation, annotation protocols vary widely across datasets -- some provide dense, frame-by-frame labels, while others rely on sparse annotations sampled at low frame rates such as 1 FPS. In this study, we investigate how such inconsistencies in annotation density and frame rate sampling influence the evaluation of zero-shot segmentation models, using SAM2 as a case study for cholecystectomy procedures. Surprisingly, we find that under conventional sparse evaluation settings, lower frame rates can appear to outperform higher ones due to a smoothing effect that conceals temporal inconsistencies. However, when assessed under real-time streaming conditions, higher frame rates yield superior segmentation stability, particularly for dynamic objects like surgical graspers. To understand how these differences align with human perception, we conducted a survey among surgeons, nurses, and machine learning engineers and found that participants consistently preferred high-FPS segmentation overlays, reinforcing the importance of evaluating every frame in real-time applications rather than relying on sparse sampling strategies. Our findings highlight the risk of evaluation bias that is introduced by inconsistent dataset protocols and bring attention to the need for temporally fair benchmarking in surgical video AI.


\end{abstract}
\section{Introduction}
Throughout history, surgical practice has been shaped by a series of transformative milestones that have fundamentally improved patient outcomes and expanded treatment possibilities~\cite{Gawande2012}. Major advances include general anesthesia~\cite{Snow1858}, antiseptic techniques~\cite{Lister1867}, and imaging modalities such as X-ray, CT, and MRI, which revolutionized surgical planning~\cite{Hounsfield1973,Lauterbur1973}. The advent of minimally invasive techniques in the 20th century substantially reduced trauma, pain, and hospital stays~\cite{Mishra2008} and more recently, the rise of computer-assisted and robotic surgery has increased surgical precision, reduced invasiveness, and enabled more complex procedures to be performed with greater accuracy~\cite{Davies2000}. Now, the next major frontier in surgery appears to be the integration of artificial intelligence (AI) at every stage of the surgical process~\cite{Esteva2021}.

Intraoperatively, AI-assisted robotic systems have the potential to enhance dexterity, accuracy, and consistency, enabling surgeons to operate with unprecedented control~\cite{Yang2017}. In particular, recent advancements in computer vision hold immense potential for real-time surgical video analysis, offering intraoperative guidance by identifying optimal dissection planes, highlighting high-risk areas, and marking safe zones for dissection~\cite{Maier-Hein2017}. Among state-of-the-art methods, video object segmentation (VOS) has recently emerged as a critical technology for surgical tool and organ segmentation~\cite{Taghavi2022}. Unlike traditional segmentation models that process frames independently, VOS models maintain spatial memory and track objects across frames, ensuring better consistency in real-time applications~\cite{Oh2019}. The SAM2 model, the state-of-the-art zero-shot segmentation framework, is particularly well-suited for surgical video analysis, as it effectively tracks and segments surgical instruments and anatomical structures without requiring task-specific fine-tuning~\cite{Kirillov2023,ravi2024sam}.

\begin{figure}[t!]
\centering
\includegraphics[width=0.24\textwidth]{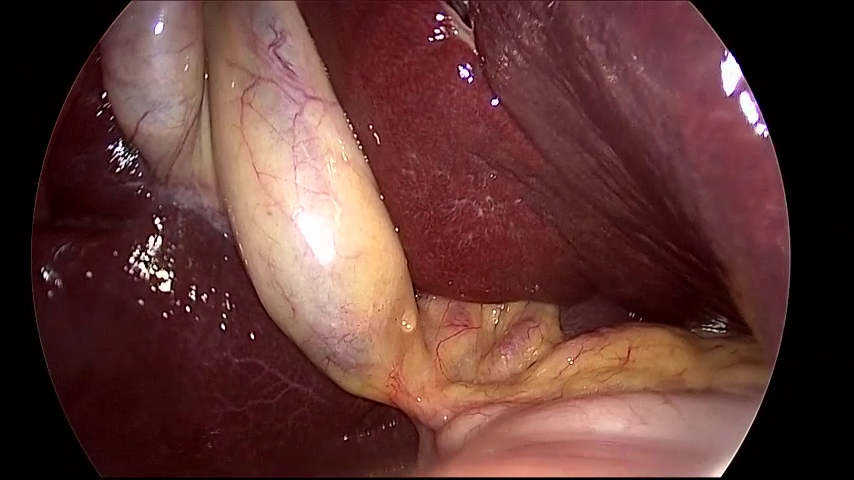}
\includegraphics[width=0.24\textwidth]{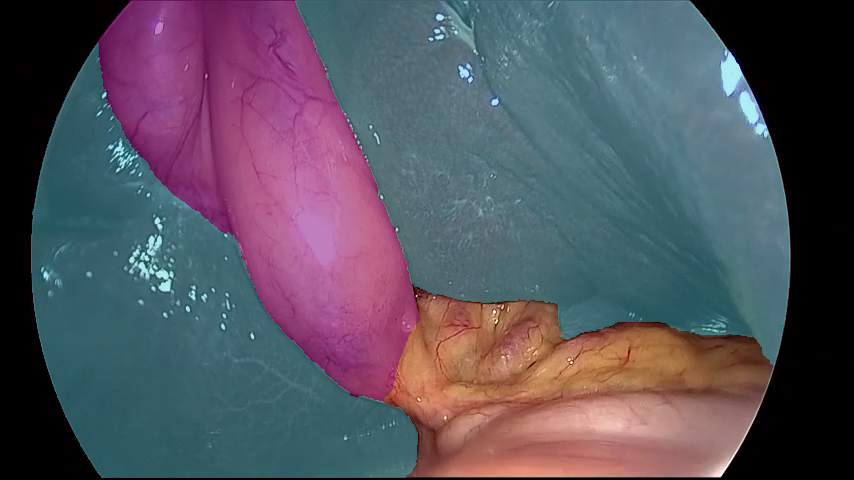}
\includegraphics[width=0.24\textwidth]{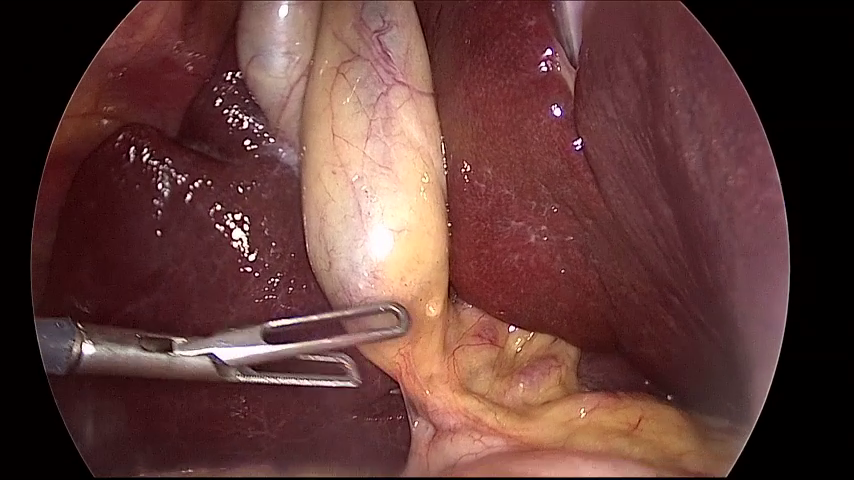}
\includegraphics[width=0.24\textwidth]{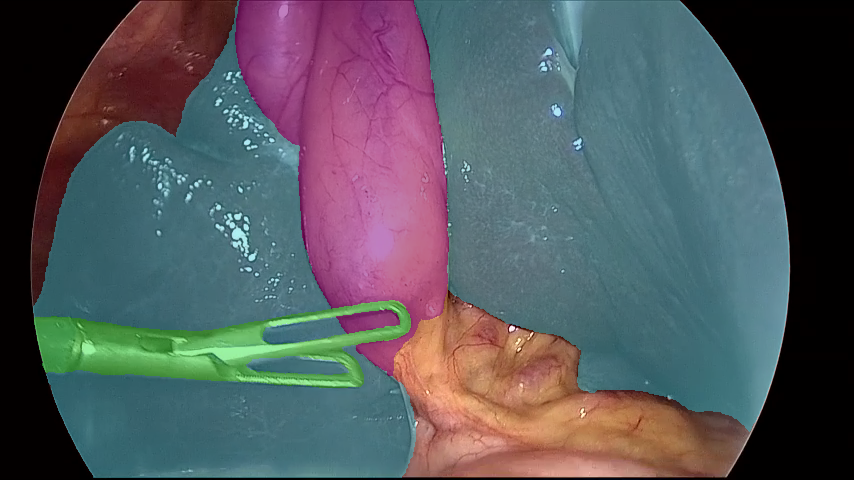}
\caption{Two example images from the CholecSeg8k dataset with annotations for the gallbladder (\gbla{\textbf{purple}}), liver (\liver{\textbf{light blue}}), and grasper (\grasper{\textbf{green}}).}
\label{fig:dataset_examples}
\end{figure}

Despite the growing interest in surgical video segmentation, annotation protocols differ substantially across datasets. For example, CholecSeg80~\cite{Czempiel2022} provides frame-wise annotations at the native recording rate of 25 FPS, whereas EndoVis 2015~\cite{Twinanda2017}, another widely used benchmark, includes labels only at a much sparser 1 FPS setting. CholecSeg8k offers densely annotated semantic masks, but only for 8,080 frames sparsely sampled across multiple videos. In contrast, the CaDIS dataset~\cite{grammatikopoulou2021cadis} includes annotations at the native 30 FPS rate. Similarly, CholecT50~\cite{alabi2025cholecinstanceseg} provides high-quality per-frame instance masks but still only at 1 FPS, limiting temporal continuity. The variation in annotation strategies and frame-rate settings is so complicated that the authors of~\cite{alabi2025cholecinstanceseg} dedicated an entire section and two figures just to clarify the annotation protocols and dataset partitions within the Cholec family. These inconsistencies in annotation density and frame sampling introduce potential evaluation biases, as segmentation performance may depend not only on model capability but also on how frequently ground truth labels are provided.

In this work, we evaluate the segmentation performance of the SAM2 model for cholecystectomy on three key targets: the gallbladder, liver, and surgical graspers. We conduct experiments across five frame-per-second (FPS) settings -- 1, 10, 15, 20, and 25 FPS -- to analyze the impact of frame rate on segmentation accuracy. Our findings reveal an intriguing phenomenon: when evaluated outside of real-time streaming scenarios, higher FPS settings may create the illusion of reduced segmentation performance. In extreme cases, our results show that the \textbf{1 FPS setting can outperform 25 FPS} when assessed under conventional evaluation methods. However, when considering real-time scenarios where predictions must remain stable across continuously arriving frames, 25 FPS setting yields superior segmentation performance, ensuring more consistent and temporally coherent segmentation. To complement these findings, we conducted a survey among surgeons, nurses, and machine learning engineers, assessing their perception of segmentation performance at different frame rates. Our findings indicate a strong preference for higher FPS segmentation mask overlays, reinforcing the importance of real-time evaluation in AI-assisted surgery.

\section{Methodology}

\subsection{Model}

In this study, we utilize SAM2.1 Hiera Large~\cite{ravi2024sam} (henceforth referred to as SAM2), a state-of-the-art zero-shot segmentation model, for real-time surgical video segmentation. SAM2 employs a powerful transformer-based architecture that ensures spatial and temporal consistency, making it particularly well suited for dynamic and complex environments such as surgical video analysis~\cite{dosovitskiy2020image,kang2024identifying,vaswani2017attention}.

Several studies have demonstrated that SAM2 outperforms both traditional image-based segmentation models and other VOS frameworks, achieving state-of-the-art results in surgical video segmentation~\cite{yu2024sam}. A key strength of SAM2 is its built-in tracking mechanism, which enhances temporal consistency across frames, reducing segmentation drift and improving overall coherence. This feature is particularly valuable in the operating room, where maintaining accurate and stable segmentation over time is crucial for real-time decision-making.

\begin{table}[t!]
\centering
\scriptsize
\caption{Total number of frames used in each video for segmentation analysis, categorized by target objects (gallbladder, liver, and surgical grasper).}
\setlength{\tabcolsep}{2.5pt}
\renewcommand{\arraystretch}{1.1}
\label{tab:data}
\begin{tabular}{lcccccc}
\toprule
Objects & Video 1 & Video 12 & Video 17 & Video 20 & Video 24 & Video 35 \\
\midrule
Any & 1,280 & 640 & 320 & 160 & 960 & 240  \\
\midrule
Gallbladder & 992 & 624 & 320 & 160 & 686 & 240 \\
Liver & 1,280 & 640 & 320 & 160 & 960 & 240  \\
Grasper & 849 & 577 & 240 & 160 & 400 & 240  \\
\bottomrule
\end{tabular}

\end{table}

\subsection{Data}

For a comprehensive evaluation across various FPS settings, we use a subset of the CholecSeg8k dataset, a frame-by-frame annotated dataset derived from the widely used Cholec80 surgical video dataset~\cite{Czempiel2022,Twinanda2017}. Cholec80 consists of 80 cholecystectomy procedure videos recorded at 25 frames-per-second at the University Hospital of Strasbourg (Strasbourg, France)~\cite{Twinanda2017}. The CholecSeg8k dataset provides high-quality segmentation annotations for a subset of those videos, making it well-suited for evaluating real-time surgical video segmentation models~\cite{Czempiel2022}.

In this study, we focus on the segmentation of two key anatomical structures -- the gallbladder and liver -- as well as a surgical instrument, the grasper. These targets are crucial for surgical scene understanding, as they provide insight into organ visibility and instrument interaction~\cite{Maier-Hein2017}. An example set of images from the dataset is provided in \figurename~\ref{fig:dataset_examples}.

Since the employed model is a zero-shot segmentation model, we do not need to partition the dataset into training and validation splits. Instead, we evaluate performance using six videos, some of which are divided into multiple segments. In total, we assess SAM2's performance across five FPS settings -- 1, 10, 15, 20, and 25 FPS -- for 15 video segments. Here, 25 FPS is the native setting in these videos and corresponds to using all available frames. Table~\ref{tab:data} shows the number of frames for each video that contain objects of each class.

\subsection{Notation and Metrics}

We represent a video as a sequence of $n$ RGB frames, denoted as $\mathsf{X} = [\texttt{X}_1, \texttt{X}_2, \ldots, \texttt{X}_n]$, where each frame \(\texttt{X} \in \mathbb{R}^{3 \times 854 \times 480}\) corresponds to an image with three color channels (RGB) and a resolution of \(854 \times 480\) pixels.

To utilize the SAM2 model for video segmentation, an initial target object must be specified in the first frame, which will then be tracked throughout the sequence. In our approach, we define this target using a binary segmentation mask, represented as $\texttt{S} \in \{0,1\}^{854 \times 480}$. This mask indicates, for each pixel, whether it belongs (\(1\)) or does not belong (\(0\)) to the target object.

Given an initial frame \(\texttt{X}_1\) and mask \(\texttt{S}_1\), SAM2 propagates segmentation across subsequent frames, predicting the \(i\)th frame as \( g(\texttt{X}_i) = \texttt{S}_i \). The mask \(\texttt{S}_i\) depends on \(\texttt{S}_1\) and all preceding frames with their inferred segmentations, ensuring temporal consistency in object tracking.

\textbf{Intersection over Union (IoU)}. In order to evaluate the correctness of predictions, we employ the IoU metric, which measures the overlap between the predicted segmentation mask \texttt{S} and the ground truth mask \texttt{Y}. Formally, IoU is defined as $\text{IoU}(\texttt{S}, \texttt{Y}) = \frac{|\texttt{S} \cap \texttt{Y}|}{|\texttt{S} \cup \texttt{Y}|}$.

\begin{table}[t!]
\centering
\scriptsize
\caption{(Evaluation: \textbf{Sampled frames}) Average IoU scores for gallbladder, liver, and surgical grasper segmentation using the SAM2 model. Results are reported for different frame rates (1, 10, 15, 20, and 25 FPS), where all available frames in each FPS setting are evaluated. The best segmentation performance for each FPS setting per row is highlighted in bold.}
\setlength{\tabcolsep}{2.2pt} 
\renewcommand{\arraystretch}{1.1} 
\label{tab:all_iou}
\begin{tabular}{lccccc|ccccc|ccccc}
\toprule
\multirow{2}{*}{\shortstack{Video\\Segment}} & \multicolumn{5}{c}{Gallbladder} & \multicolumn{5}{c}{Liver} & \multicolumn{5}{c}{Grasper} \\
\cmidrule[0.5pt]{2-16}
~ \phantom{---}& 1 & 10 & 15 & 20 & 25 & 1 & 10 & 15 & 20 & 25 & 1 & 10 & 15 & 20 & 25 \\
\midrule
V1 S1 & \textbf{96.0} & 95.7 & 95.9 & 95.9 & 95.9 & \textbf{90.9} & 90.3 & 90.4 & 90.2 & 90.3 & \textbf{87.8} & 86.3 & 86.4 & 86.5 & 86.3 \\
V1 S2 & \textbf{96.4} & 96.1 & 96.3 & 96.2 & 96.1 & \textbf{97.2} & 96.6 & 97.0 & 97.1 & 96.9 & \textbf{91.2} & 86.8 & 86.7 & 87.0 & 86.9 \\
V1 S3 & \textbf{95.4} & 94.9 & 94.9 & 94.9 & 94.9 & \textbf{96.3} & 96.0 & 96.1 & 96.2 & 96.2 & \textbf{83.9} & 82.1 & 82.2 & 82.4 & 82.1 \\
V1 S4 & \textbf{86.5} & 85.5 & 85.1 & 84.9 & 85.2 & \textbf{96.0} & 95.6 & 95.6 & 95.6 & 95.6 & \textbf{82.3} & 80.3 & 80.2 & 79.9 & 80.2 \\
V1 S5 & \textbf{87.8} & 86.7 & 87.5 & 86.7 & 86.8 & \textbf{93.5} & 92.3 & 92.2 & 92.2 & 92.2 & 78.2 & 80.4 & 80.5 & 80.6 & \textbf{80.7} \\
\midrule
V12 S1 & 75.4 & \textbf{82.4} & 81.7 & 81.7 & 81.6 & \textbf{92.6} & 91.7 & 91.1 & 91.2 & 91.1 & 78.9 & \textbf{86.5} & 86.4 & 86.5 & 86.5 \\
V12 S2 & \textbf{71.5} & 67.0 & 67.4 & 68.7 & 67.7 & \textbf{94.0} & 91.9 & 91.1 & 91.4 & 91.4 & 54.2 & \textbf{54.4} & 53.7 & 53.9 & 53.9 \\
V12 S3 & \textbf{81.2} & 78.2 & 78.5 & 77.6 & 77.9 & 84.6 & \textbf{88.5} & 88.1 & 88.3 & 88.0 & 55.2 & \textbf{55.4} & 55.3 & 55.2 & 55.0 \\
\midrule
V17 S1 & \textbf{95.9} & 95.5 & 94.8 & 94.9 & 95.0 & \textbf{95.0} & 93.8 & 92.6 & 93.1 & 93.0 & \textbf{74.5} & 69.8 & 69.7 & 69.3 & 69.4 \\
V17 S2 & \textbf{96.3} & 95.1 & 95.2 & 95.1 & 95.2 & 94.1 & 92.7 & \textbf{93.4} & 93.1 & 93.2 & \textbf{88.5} & 86.9 & 86.7 & 86.1 & 86.1 \\
\midrule
V20 S1 & \textbf{93.5} & 93.1 & 93.3 & 93.2 & 93.0 & \textbf{96.7} & 96.4 & 96.4 & 96.4 & 96.4 & \textbf{95.1} & 94.6 & 94.6 & 94.5 & 94.5 \\
\midrule
V24 S4 & \textbf{90.2} & 81.4 & 80.6 & 78.4 & 78.3 & 94.1 & \textbf{95.4} & 95.4 & 95.4 & 95.3 & \textbf{88.0} & 87.1 & 87.2 & 86.9 & 86.9 \\
\midrule
V35 S1 & \textbf{95.1} & 93.7 & 93.8 & 93.8 & 93.7 & \textbf{98.0} & 97.3 & 97.5 & 97.6 & 97.6 & \textbf{96.2} & 94.9 & 94.1 & 94.1 & 94.3 \\
V35 S2 & \textbf{95.9} & 95.6 & 94.9 & 94.9 & 95.0 & 97.4 & \textbf{97.8} & 97.8 & 97.8 & 97.8 & \textbf{95.1} & 93.1 & 93.3 & 93.2 & 93.3 \\
V35 S3 & \textbf{93.9} & 92.1 & 92.0 & 92.0 & 92.0 & \textbf{98.3} & 98.2 & 98.2 & 98.2 & 98.2 & \textbf{97.0} & 96.3 & 96.2 & 96.2 & 96.2 \\
\bottomrule
\end{tabular}

\end{table}

\subsection{Evaluation}
\label{sec:evaluation}

In this work, we evaluate the performance of SAM2 on surgical videos sampled in three settings: sampled frames, anchor frames, and real-time streaming frames.

\textbf{Sampled frames}. In this setting, we evaluate the IoU over sampled frames within the selected FPS setting. Formally, given a frame sequence $\mathsf{X}$ captured at 25 FPS, the evaluation is conducted over the subset $\mathsf{X}^{(f)} = [\texttt{X}_1, \texttt{X}_{1+\Delta f}, \texttt{X}_{1+2\Delta f}, \dots]$, where $\Delta f = 25/f$ determines the interval between selected frames. This method ensures that higher FPS settings use more frames for evaluation, but it introduces potential bias since different FPS settings involve different numbers of frames. For example, in a 10-second video recorded at 25 FPS, there are $10 \times 25 = 250$ total frames, whereas at 1 FPS, there are only $10 \times 1 = 10$ frames. This means that higher FPS settings provide more temporal details, which can influence the evaluation results. To mitigate this bias, we introduce the anchor frames setting.

\textbf{Anchor frames}. In this setting, we standardize the number of frames used for evaluation across different FPS settings. Specifically, we define the anchor frame set as $\mathsf{X}_{\text{Anchor}} = [\texttt{X}_1, \texttt{X}_{26}, \texttt{X}_{51}, \dots]$, where only the first frame of each second is used, ensuring that all FPS settings are evaluated on the same number of frames. This approach eliminates potential biases introduced by varying frame counts and provides a fairer comparison across different FPS configurations.

\textbf{Real-time streaming frames}. This setting simulates a real-time streaming scenario where predictions from lower FPS settings must persist across intermediate frames. Regardless of the selected FPS setting, evaluation is conducted on all 25 frames per second. Formally, for an FPS $f$, predictions $\texttt{S}_t$ at sampled frames are propagated forward across intermediate frames until the next sampled frame appears. That is, for an FPS setting of $f$, a prediction made at $\texttt{X}_t$ remains unchanged for frames $\texttt{X}_{t+1}, \dots, \texttt{X}_{t+\Delta f -1}$ until the next prediction update occurs at $\texttt{X}_{t+\Delta f}$. This setup assesses the practical impact of frame rate on segmentation performance in real-time applications, where predictions must remain valid between updates to prevent flickering.

\begin{table}[t!]
\centering
\scriptsize
\caption{(Evaluation: \textbf{Anchor frames}) Average IoU scores for gallbladder, liver, and surgical grasper segmentation using the SAM2 model when evaluated only on anchor frames (the first frame of each second). The best segmentation performance for each FPS setting per row is highlighted in bold.}
\setlength{\tabcolsep}{2.2pt} 
\renewcommand{\arraystretch}{1.1} 
\label{tab:anchor_iou}
\begin{tabular}{lccccc|ccccc|ccccc}
\toprule
\multirow{2}{*}{\shortstack{Video\\Segment}} & \multicolumn{5}{c}{Gallbladder} & \multicolumn{5}{c}{Liver} & \multicolumn{5}{c}{Grasper} \\
\cmidrule[0.5pt]{2-16}
~ \phantom{---}& 1 & 10 & 15 & 20 & 25 & 1 & 10 & 15 & 20 & 25 & 1 & 10 & 15 & 20 & 25 \\
\midrule
V1 S1 & 96.3 & 96.4 & 96.4 & \textbf{96.5} & \textbf{96.5} & 90.9 & \textbf{91.1} & 91.0 & 90.9 & 90.6 & 87.8 & 87.9 & \textbf{88.1} & 88.0 & 87.9 \\
V1 S2 & \textbf{97.2} & \textbf{97.2} & \textbf{97.2} & \textbf{97.2} & \textbf{97.2} & \textbf{97.2} & \textbf{97.2} & \textbf{97.2} & \textbf{97.2} & \textbf{97.2} & \textbf{91.2} & 90.5 & 90.2 & 90.5 & 90.6 \\
V1 S3 & 95.4 & \textbf{95.6} & \textbf{95.6} & 95.5 & 95.4 & 96.3 & 96.4 & 96.4 & \textbf{96.5} & \textbf{96.5} & \textbf{96.2} & \textbf{96.2} & \textbf{96.2} & \textbf{96.2} & \textbf{96.2} \\
V1 S4 & 94.1 & 95.7 & \textbf{95.8} & \textbf{95.8} & 95.7 & 96.0 & 95.9 & \textbf{96.0} & 95.9 & 95.9 & 75.4 & 77.8 & 77.5 & 77.7 & \textbf{78.1} \\
V1 S5 & 93.5 & 93.3 & \textbf{93.4} & \textbf{93.4} & \textbf{93.4} & 93.5 & 93.3 & \textbf{93.4} & \textbf{93.4} & \textbf{93.4} & 82.3 & \textbf{82.4} & 82.3 & 82.1 & 82.1 \\
\midrule
V12 S1 & \textbf{95.0} & \textbf{95.0} & 94.9 & 94.9 & \textbf{95.0} & 92.6 & \textbf{92.8} & 92.2 & 92.2 & 92.2 & 83.5 & 83.4 & \textbf{83.5} & 83.4 & 83.2 \\
V12 S2 & \textbf{96.7} & \textbf{96.7} & 96.6 & 96.6 & \textbf{96.7} & 94.0 & \textbf{94.1} & 94.0 & 94.0 & 93.9 & \textbf{95.0} & \textbf{95.0} & 94.9 & 94.9 & \textbf{95.0} \\
V12 S3 & 94.1 & 94.0 & 94.0 & 93.9 & \textbf{94.1} & 84.6 & 88.8 & 87.0 & \textbf{88.9} & 86.9 & \textbf{89.4} & 89.3 & 89.3 & \textbf{89.4} & 89.1 \\
\midrule
V17 S1 & \textbf{95.9} & 95.8 & 95.8 & 95.8 & \textbf{95.9} & 95.0 & \textbf{95.1} & \textbf{95.1} & \textbf{95.1} & 93.8 & 95.2 & \textbf{95.3} & \textbf{95.3} & 95.2 & 95.2 \\
V17 S2 & 96.0 & 96.0 & \textbf{96.1} & \textbf{96.1} & 96.0 & \textbf{94.1} & \textbf{94.1} & \textbf{94.1} & \textbf{94.1} & \textbf{94.1} & \textbf{95.3} & \textbf{95.3} & \textbf{95.3} & 95.2 & 95.1 \\
\midrule
V20 S1 & \textbf{96.8} & \textbf{96.8} & \textbf{96.8} & \textbf{96.8} & \textbf{96.8} & 96.7 & \textbf{96.8} & \textbf{96.8} & \textbf{96.8} & \textbf{96.8} & \textbf{95.3} & \textbf{95.3} & \textbf{95.3} & 95.2 & 95.1 \\
\midrule
V24 S4 & \textbf{90.2} & 80.2 & 79.9 & 76.1 & 76.7 & 94.1 & 95.7 & \textbf{95.8} & \textbf{95.8} & 95.7 & 95.7 & \textbf{95.8} & \textbf{95.8} & 95.7 & 95.7 \\
\midrule
V35 S1 & 98.3 & \textbf{98.5} & \textbf{98.5} & \textbf{98.5} & \textbf{98.5} & 98.0 & \textbf{98.2} & \textbf{98.2} & \textbf{98.2} & \textbf{98.2} & 96.8 & 96.8 & 96.8 & 96.8 & \textbf{97.0} \\
V35 S2 & 98.0 & \textbf{98.0} & \textbf{98.0} & \textbf{98.0} & \textbf{98.0} & 97.4 & \textbf{98.0} & \textbf{98.0} & \textbf{98.0} & \textbf{98.0} & 95.2 & \textbf{95.3} & \textbf{95.3} & 95.2 & 95.1 \\
V35 S3 & 97.4 & \textbf{98.0} & \textbf{98.0} & \textbf{98.0} & \textbf{98.0} & 98.3 & \textbf{98.5} & \textbf{98.5} & \textbf{98.5} & \textbf{98.5} & 96.8 & 96.8 & 96.8 & 96.8 & \textbf{97.0} \\
\bottomrule
\end{tabular}

\end{table}

\section{Experimental Results}

In this section, we present the evaluation results of the SAM2 model across the three evaluation strategies highlighted in Section~\ref{sec:evaluation}: (1) sampled frames, (2) anchor frames, and (3) real-time streaming frames. Results for these settings can be found in Table~\ref{tab:all_iou}, Table~\ref{tab:anchor_iou}, and Table~\ref{tab:real_time_iou}, respectively.

\textbf{Sampled frames}. This setting assesses segmentation accuracy by computing the IoU for the sampled frames available in a given FPS setting. Surprisingly, in most cases, the 1 FPS setting outperforms the 25 FPS setting (see Table~\ref{tab:all_iou}). This counterintuitive result arises because lower FPS evaluates performance on significantly fewer frames. By selecting only one frame per second, this setting inherently smooths out segmentation inconsistencies, making the results appear more stable and artificially inflating IoU scores. Higher FPS settings, on the other hand, introduce more frequent updates, which expose even minor segmentation variations. These variations, while not necessarily indicative of poor performance, result in a slight reduction in IoU scores. Our observations suggest that this evaluation setting may not be the most reliable method for assessing real-world segmentation stability, as it can mask critical segmentation inconsistencies during operations that become apparent at higher frame rates.

\begin{table}[t!]
\centering
\scriptsize
\caption{(Evaluation: \textbf{Real-time streaming}) Average IoU scores for gallbladder, liver, and surgical grasper segmentation using the SAM2 model in a real-time streaming scenario, where lower FPS predictions persist across intermediate frames. The best segmentation performance for each FPS setting per row is highlighted in bold.}
\setlength{\tabcolsep}{2.2pt} 
\renewcommand{\arraystretch}{1.1} 
\label{tab:real_time_iou}
\begin{tabular}{lccccc|ccccc|ccccc}
\toprule
\multirow{2}{*}{\shortstack{Video\\Segment}} & \multicolumn{5}{c}{Gallbladder} & \multicolumn{5}{c}{Liver} & \multicolumn{5}{c}{Grasper} \\
\cmidrule[0.5pt]{2-16}
~ \phantom{---}& 1 & 10 & 15 & 20 & 25 & 1 & 10 & 15 & 20 & 25 & 1 & 10 & 15 & 20 & 25 \\
\midrule
V1 S1 & 79.8 & 94.1 & 95.1 & 95.5 & \textbf{95.9} & 81.8 & 89.4 & 89.9 & 89.9 & \textbf{90.3} & 30.8 & 35.9 & 39.7 & 42.2 & \textbf{74.4} \\
V1 S2 & 88.5 & 95.5 & 95.8 & 96.0 & \textbf{96.1} & 90.9 & 95.9 & 96.4 & 96.6 & \textbf{96.9} & 57.5 & 83.3 & 84.6 & 85.9 & \textbf{86.9} \\
V1 S3 & 84.2 & 94.0 & 94.5 & 94.7 & \textbf{94.9} & 83.3 & 94.8 & 95.5 & 95.9 & \textbf{96.2} & 63.5 & 84.4 & 85.5 & 86.0 & \textbf{86.3} \\
V1 S4 & 67.7 & 82.7 & 83.9 & 84.2 & \textbf{85.2} & 79.9 & 94.0 & 94.8 & 95.2 & \textbf{95.6} & 37.7 & 73.5 & 77.0 & 79.4 & \textbf{82.1} \\
V1 S5 & 61.8 & 84.1 & 86.3 & 86.2 & \textbf{86.8} & 81.3 & 91.5 & 91.9 & 92.0 & \textbf{92.2} & 37.4 & 75.4 & 77.8 & 79.2 & \textbf{80.2} \\
\midrule
V12 S1 & 62.1 & 80.0 & 81.6 & 81.5 & \textbf{81.6} & 79.1 & 90.5 & 90.8 & 91.0 & \textbf{91.1} & 34.8 & 73.9 & 77.8 & 79.2 & \textbf{80.7} \\
V12 S2 & 57.8 & 67.0 & 67.0 & \textbf{68.8} & 67.7 & 83.7 & 90.8 & 91.2 & 91.4 & \textbf{91.4} & 41.5 & 52.6 & 53.1 & 53.8 & \textbf{53.9} \\
V12 S3 & 59.5 & 76.9 & 77.4 & 77.8 & \textbf{77.9} & 70.4 & 86.8 & 87.5 & 87.8 & \textbf{88.0} & 38.7 & 80.5 & 83.5 & 84.8 & \textbf{86.5} \\
\midrule
V17 S1 & 66.2 & 92.0 & 93.5 & 94.4 & \textbf{95.0} & 69.3 & 90.4 & 91.6 & 92.4 & \textbf{93.0} & 30.9 & 64.0 & 66.4 & 67.7 & \textbf{69.4} \\
V17 S2 & 78.6 & 93.0 & 94.1 & 94.6 & \textbf{95.2} & 74.8 & 91.4 & 92.3 & 92.8 & \textbf{93.2} & 34.7 & 75.9 & 80.1 & 83.3 & \textbf{86.1} \\
\midrule
V20 S1 & 83.5 & 92.1 & 92.7 & 93.0 & \textbf{93.0} & 89.4 & 95.9 & 96.2 & 96.3 & \textbf{96.4} & 79.2 & 92.7 & 93.7 & 94.2 & \textbf{94.5} \\
\midrule
V24 S4 & 45.1 & 81.7 & 83.6 & 85.0 & \textbf{86.3} & 74.2 & 90.2 & 91.3 & 91.9 & \textbf{92.4} & \textbf{86.9} & 83.2 & 80.6 & 75.9 & 31.2 \\
\midrule
V35 S1 & 56.3 & 85.6 & 89.9 & 91.8 & \textbf{93.7} & 83.6 & 95.7 & 96.7 & 97.1 & \textbf{97.6} & 29.8 & 77.3 & 85.6 & 90.4 & \textbf{94.3} \\
V35 S2 & 56.2 & 90.3 & 92.7 & 93.9 & \textbf{95.0} & 73.6 & 95.5 & 96.9 & 97.5 & \textbf{97.8} & 25.2 & 78.8 & 86.5 & 89.9 & \textbf{93.3} \\
V35 S3 & 69.0 & 89.4 & 90.7 & 91.3 & \textbf{92.0} & 86.6 & 97.2 & 97.7 & 97.9 & \textbf{98.2} & 45.8 & 87.8 & 91.7 & 94.1 & \textbf{96.2} \\
\bottomrule
\end{tabular}
\end{table}

\textbf{Anchor frames}. To address potential biases introduced by varying frame counts across FPS settings, the anchor frames setting evaluates segmentation only on the first frame of each second. As shown in Table~\ref{tab:anchor_iou}, this setting reveals no apparent superior FPS choice, with segmentation performance remaining relatively similar across all frame rates.

\textbf{Real-time streaming frames}. This setting simulates real-world deployment by requiring lower FPS predictions to persist across intermediate frames to prevent flickering. This evaluation reflects practical constraints in real-time applications, where model outputs must remain stable between updates. As illustrated in Table~\ref{tab:real_time_iou}, segmentation accuracy consistently favors 25 FPS, particularly for the surgical grasper, which is the object that moves by far the most in surgical operations. Unlike traditional evaluation settings, real-time evaluation clearly demonstrates that higher FPS settings are crucial for maintaining segmentation consistency and preventing temporal drift.

\subsection{Human Perception of Annotation FPS in Surgical Videos}

Following our experiments, we conducted a survey to evaluate the human perception of segmentation frame rates in surgical videos. We asked participants, including three surgeons, three nurses, and four machine learning engineers, to compare videos with segmentation maps generated at different frame rates (e.g., 25v22, 25v20, 25v15) and indicate whether they preferred the higher frame rate, the lower frame rate, or had no preference. Participants were only provided with the segmented video, and did not know the underlying segmentation frame rate.
Our goal in this survey is to assess how variations in segmentation FPS impact perceived quality and whether lower FPS affects visual perception.

\begin{figure}[t!]
\centering
\includegraphics[width=0.98\textwidth]{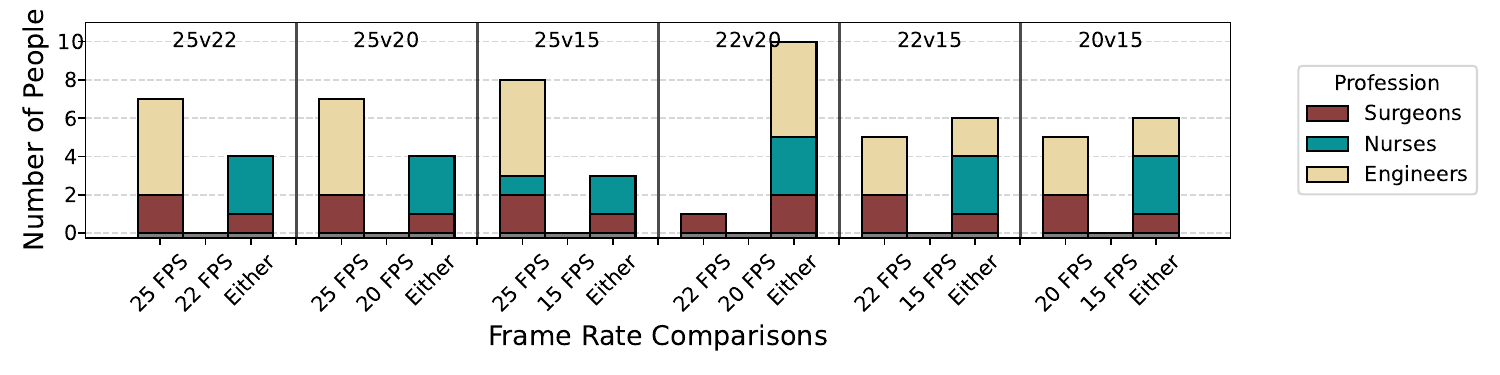}
\caption{Illustration of how surgeons, nurses, and engineers perceive differing segmentation frame rates in 25 FPS surgical videos. Each grouped set of bars represents a frame rate comparison between two videos, with responses categorized into three choices: preferring the first video/FPS (left bar), preferring the second video/FPS (middle bar), or no preference/either (right bar).}
\label{fig:survey}
\end{figure}

The survey results provided in Figure~\ref{fig:survey} indicate that no participants preferred segmentation maps generated at lower frame rates in any comparison. Instead, when differences were subtle, respondents tended to select the ``either is okay'' option. As the gap in segmentation FPS increased, more participants noticed a difference, with a preference for higher FPS. Most participants described segmentation mask overlays with lower FPS settings as being \say{choppy} and out-of-sync.

\section{Conclusions}


In this work, we revisited the impact of frame rate sampling on zero-shot surgical video segmentation through a fairness lens, focusing on how evaluation strategies can introduce unintended biases in model performance assessment. Using the SAM2 model, we evaluated segmentation accuracy across multiple frame sampling rates and found a counterintuitive result: under conventional evaluation settings, lower frame rates often appear to outperform higher ones. This apparent advantage is largely due to a smoothing effect -- fewer segmentation updates result in fewer opportunities for visible errors. However, when models are evaluated under real-time streaming conditions, where predictions must remain temporally consistent across all frames, higher frame rates yield better performance by maintaining segmentation stability and coherence.

To understand how these effects align with human perception, we conducted a survey involving surgeons, nurses, and machine learning engineers. The results highlighted a clear bias in perceptual preference: as the time gap between segmentation mask overlays increased, participants consistently favored higher FPS outputs. This suggests that evaluation protocols based solely on sparse-frame metrics may obscure issues critical to clinical usability and reinforces the importance of temporally aware evaluation in AI-assisted surgery. As such, we call for fairer evaluation practices in surgical video segmentation that move beyond sparse-frame sampling and better reflect both clinical expectations and temporal stability.

\bibliographystyle{splncs04}
\bibliography{FAIMI_30}

\begin{thebibliography}{10}
\providecommand{\url}[1]{\texttt{#1}}
\providecommand{\urlprefix}{URL }
\providecommand{\doi}[1]{https://doi.org/#1}

\bibitem{alabi2025cholecinstanceseg}
Alabi, O., Toe, K.K.Z., Zhou, Z., Budd, C., Raison, N., Shi, M., Vercauteren, T.: Cholecinstanceseg: A tool instance segmentation dataset for laparoscopic surgery. Scientific Data  \textbf{12}(1),  1--12 (2025)

\bibitem{Davies2000}
Davies, B.: A review of robotics in surgery. Proceedings of the Institution of Mechanical Engineers, Part H: Journal of Engineering in Medicine  \textbf{214}(1),  129--140 (2000)

\bibitem{dosovitskiy2020image}
Dosovitskiy, A., Beyer, L., Kolesnikov, A., Weissenborn, D., Zhai, X., Unterthiner, T., Dehghani, M., Minderer, M., Heigold, G., Gelly, S., et~al.: An image is worth 16x16 words: Transformers for image recognition at scale. arXiv preprint arXiv:2010.11929  (2020)

\bibitem{Esteva2021}
Esteva, A., Chou, K., Yeung, S., Naik, N., Madani, A., Mottaghi, A.: Deep learning-enabled medical computer vision. npj Digital Medicine  \textbf{4}(5) (2021)

\bibitem{Gawande2012}
Gawande, A.: Better: A Surgeon's Notes on Performance. Picador (2012)

\bibitem{grammatikopoulou2021cadis}
Grammatikopoulou, M., Flouty, E., Kadkhodamohammadi, A., Quellec, G., Chow, A., Nehme, J., Luengo, I., Stoyanov, D.: Cadis: Cataract dataset for surgical rgb-image segmentation. Medical Image Analysis  \textbf{71},  102053 (2021)

\bibitem{Czempiel2022}
Hong, W.Y., Kao, C.L., Kuo, Y.H., Wang, J.R., Chang, W.L., Shih, C.S.: Cholecseg8k: a semantic segmentation dataset for laparoscopic cholecystectomy based on cholec80. arXiv preprint arXiv:2012.12453  (2020)

\bibitem{Hounsfield1973}
Hounsfield, G.N.: Computerized transverse axial scanning (tomography): Part 1. description of system. British Journal of Radiology  \textbf{46},  1016--1022 (1973)

\bibitem{kang2024identifying}
Kang, S., Vankerschaver, J., Ozbulak, U.: Identifying critical tokens for accurate predictions in transformer-based medical imaging models. In: International Workshop on Machine Learning in Medical Imaging. pp. 169--179. Springer (2024)

\bibitem{Kirillov2023}
Kirillov, A., Mintun, E., Ravi, N., et~al.: Segment anything. arXiv preprint  \textbf{arXiv:2304.02643} (2023)

\bibitem{Lauterbur1973}
Lauterbur, P.C.: Image formation by induced local interactions: Examples employing nuclear magnetic resonance. Nature  \textbf{242},  190--191 (1973)

\bibitem{Lister1867}
Lister, J.: On the antiseptic principle in the practice of surgery. The Lancet  \textbf{90}(2299),  353--356 (1867)

\bibitem{Maier-Hein2017}
Maier-Hein, L., Engelhardt, S., Syben, A.M.R., et~al.: Computer-assisted medical image analysis for intervention planning: Current and future challenges. Medical Image Analysis  \textbf{33},  66--75 (2017)

\bibitem{Mishra2008}
Mishra, R.: Minimally Invasive Surgery. Jaypee Brothers Medical Publishers (2008)

\bibitem{Oh2019}
Oh, S.W., Lee, J.Y., Xu, N., Kim, S.J.: Video object segmentation using space-time memory networks. IEEE Conference on Computer Vision and Pattern Recognition (CVPR) pp. 9226--9235 (2019)

\bibitem{ravi2024sam}
Ravi, N., Gabeur, V., Hu, Y.T., Hu, R., Ryali, C., Ma, T., Khedr, H., R{\"a}dle, R., Rolland, C., Gustafson, L., et~al.: Sam 2: Segment anything in images and videos. arXiv preprint arXiv:2408.00714  (2024)

\bibitem{Snow1858}
Snow, J.: On chloroform and other anaesthetics: their action and administration (1858)

\bibitem{Taghavi2022}
Taghavi, N., Fereshtehnejad, S.M., Navab, N.: Advancements in deep learning-based video object segmentation for medical applications. IEEE Transactions on Medical Imaging  \textbf{41}(12),  3152--3166 (2022)

\bibitem{Twinanda2017}
Twinanda, A., Shehata, S., Mutter, D., Marescaux, J., de~Mathelin, M., Padoy, N.: Endonet: A deep architecture for recognition tasks on laparoscopic videos. IEEE Transactions on Medical Imaging  \textbf{36}(1),  86--97 (2017)

\bibitem{vaswani2017attention}
Vaswani, A., Shazeer, N., Parmar, N., Uszkoreit, J., Jones, L., Gomez, A.N., Kaiser, {\L}., Polosukhin, I.: Attention is all you need. Advances in neural information processing systems  \textbf{30} (2017)

\bibitem{Yang2017}
Yang, G.Z., Tsang, J.J.C., Martin, J.: From autonomous systems to autonomous robots: The next medical revolution? Proceedings of the IEEE  \textbf{105}(10),  1954--1966 (2017)

\bibitem{yu2024sam}
Yu, J., Wang, A., Dong, W., Xu, M., Islam, M., Wang, J., Bai, L., Ren, H.: Sam 2 in robotic surgery: An empirical evaluation for robustness and generalization in surgical video segmentation. arXiv preprint arXiv:2408.04593  (2024)

\end{thebibliography}

\end{document}